\documentclass{article}

\usepackage{PRIMEarxiv}

\usepackage[utf8]{inputenc} 
\usepackage[T1]{fontenc}    
\usepackage{hyperref}       
\usepackage{url}            
\usepackage{booktabs}       
\usepackage{amsfonts}       
\usepackage{nicefrac}       
\usepackage{microtype}      
\usepackage{lipsum}
\usepackage{fancyhdr}       
\usepackage{graphicx}       
\graphicspath{{media/}}     
\usepackage{xcolor,soul,framed} 
\usepackage{subcaption}
\usepackage{amsmath}
\usepackage{algorithmicx}
\usepackage{algorithm}
\usepackage{algpseudocode}

\title{Detecting misinformation through Framing Theory: the Frame Element-based Model
}

\author{
  Guan Wang \\
  Auckland University of Technology \\
  Auckland, New Zealand\\
  \texttt{guan.wang@autuni.ac.nz} \\
   \And
  Rebecca Frederick \\
  Auckland University of Technology \\
  Auckland, New Zealand\\
  \texttt{ngt8261@autuni.ac.nz} \\
   \And
   Jinglong Duan \\
  Auckland University of Technology \\
  Auckland, New Zealand\\
  \texttt{jinglong.duan@autuni.ac.nz} \\
   \And
   William Wong \\
  Auckland University of Technology \\
  Auckland, New Zealand\\
  \texttt{william.wong@aut.ac.nz} \\
   \And
   Verica Rupar \\
  Auckland University of Technology \\
  Auckland, New Zealand\\
  \texttt{verica.rupar@aut.ac.nz} \\
   \And
   Weihua Li \\
  Auckland University of Technology \\
  Auckland, New Zealand\\
  \texttt{weihua.li@aut.ac.nz} \\
   \And
  Quan Bai \\
  University of Tasmania \\
  Hobart, Australia\\
  \texttt{quan.bai@utas.edu.au} \\
}

\begin{document}
\maketitle

\begin{abstract}
In this paper, we delve into the rapidly evolving challenge of misinformation detection, with a specific focus on the nuanced manipulation of narrative frames — an under-explored area within the AI community. The potential for Generative AI models to generate misleading narratives underscores the urgency of this problem. Drawing from communication and framing theories, we posit that the presentation or `framing' of accurate information can dramatically alter its interpretation, potentially leading to misinformation. We highlight this issue through real-world examples, demonstrating how shifts in narrative frames can transmute fact-based information into misinformation. To tackle this challenge, we propose an innovative approach leveraging the power of pre-trained Large Language Models and deep neural networks to detect misinformation originating from accurate facts portrayed under different frames. These advanced AI techniques offer unprecedented capabilities in identifying complex patterns within unstructured data critical for examining the subtleties of narrative frames. The objective of this paper is to bridge a significant research gap in the AI domain, providing valuable insights and methodologies for tackling framing-induced misinformation, thus contributing to the advancement of responsible and trustworthy AI technologies. Several experiments are intensively conducted and experimental results explicitly demonstrate the various impact of elements of framing theory proving the rationale of applying framing theory to increase the performance in misinformation detection.
\end{abstract}


\section{Introduction}

Misinformation in today's media landscape is growing substantively, where fake news and false or misleading information is disseminated through various media channels, e.g., news stories and online social media platforms \cite{islam2020deep}. Artificial Intelligence (AI) has advanced from exclusively understanding language to Generative AI (GAI) models that can automatically generate articles, posts, and narratives with remarkable sophistication \cite{longoni2022news}. The accessibility of GAI models such as ChatGPT has expedited the process of creating manipulative misinformation, and in most cases, it can be difficult for readers to distinguish whether the narrative was written by a GAI model or a human author \cite{kshetri2023chatgpt}. Automating fact-checking or claim validation is a well-researched task that has achieved high accuracy results with traditional misinformation detection focused on keywords \cite{rashkin2017truth,tchechmedjiev2019claimskg}. However, when presented with accurate facts where frames have been used to manipulate the information of the narrative to be misleading, it is difficult to identify the misinformation. The framing of accurate information by selecting and highlighting some aspects while simultaneously omitting other aspects to present an alternative perspective can lead to the communication of a different message than the original, accurate narrative intended \cite{entman2004projections}. The aforementioned manipulation of accurate information by changing the perspective and frame can result in the propagation of misinformation, thus, framing plays an important role in misinformation detection.

Framing theory illuminates the process by which communicators strategically highlight specific facets of a perceived reality within a communication text \cite{entman1993framing}. This intentional emphasis serves to advance a distinct problem definition, causal interpretation, moral evaluation, and/or treatment recommendation \cite{entman1993framing}. Framing involves the selection of some factors about an issue or event, and to make salient, or to emphasize, these factors over other factors. It is about selecting and deciding which parts of a situation or event to make salient to an audience. It also suggests how information is presented and communicated in a narrative - the story that communicates the facts in a meaningful way - can influence an individual's perception and interpretation of that information and is recognized as an important concept in the communication and social science fields \cite{entman2004projections,entman1993framing,goffman1974frame,scheufele1999framing,fairhurst1996art}. Additionally, framing theory suggests that four frame elements contribute to how information is presented: problem definition, causal interpretation, moral evaluation, and treatment recommendation \cite{entman1993framing}. Specifically, the problem definition defines the problem by determining the actions of a causal agent along with their associated costs and benefits and is measured by what is culturally acceptable, while the causal interpretation identifies which forces cause the problem. The moral evaluation makes moral judgments on the causal agent and their effects, with the treatment recommendation offering suggestions to solve the problem and the possible effects these might have. When it comes to misinformation, framing theory suggests that the manner in which information is conveyed or framed can be harnessed to persuade readers into embracing inaccurate or misleading information as truthful. By strategically highlighting specific facts or interpretations while purposefully excluding others, individuals or organizations can craft a specific narrative that aligns with their agenda to mislead the audience \cite{entman2004projections}.


There are numerous research studies concerning the detection of framing, the identification of elements within a frame, and the analysis of framing itself \cite{entman1993framing,walter2019news,touri2015using}. However, few studies explore how frames impact the emergence of misinformation or which frame element has the greatest impact on the overall frame. It is challenging to classify misinformation stemming from factual information. Therefore, misleading information created by manipulating the frame of a truthful narrative would be undetected by traditional misinformation detection models. In this study, we used three topics that stirred public controversy to evaluate the detection model we were developing: the Three Waters Reform media debate in New Zealand, coverage of Covid-19 globally, and reporting on Nuclear Pollution – more on data sets later in the texts. For example, an excerpt of a factual information narrative with a political frame:

\textit{``The proposed three waters reform program harks back to the Havelock North water contamination event in 2016... The government estimates that we'll need a mind-boggling \$120 billion to \$185 billion over the next 30 years... The government believes that four entities, aggregating all the water services across the country, offer the best and quickest opportunity to achieve the desired improvements... The review was expanded to cover all three waters, this acknowledges the inter-relationships between the three networks."} 

Information is presented in a straightforward and factual manner, explaining the motivation behind the reform, the expected costs, and the time frame, describing the government's belief that larger entities can achieve efficiency gains, and understanding why all three water networks were reviewed. However, an excerpt of a misleading narrative with a semantic frame that uses specific terms to associate the statement with other communication contents or features, including irony, lettering, metaphor and so on \cite{SCHEIBENZUBER2023107587}. The following example shows the satire/irony which suggests the opposite of the original message:

\textit{``Because nothing says `clean water' like shifting responsibility from local government to some fancy-sounding entity, right?... They even established a drinking water regulator to ensure everything meets regulatory standards, because we all know how important it is to regulate things, right?... Because who needs small, local councils when you can have these big entities making all the decisions for you? Efficiency gains are just a bonus, my friends!... Because why bother keeping it simple when you can add some unnecessary complexity?"}

Satire, oversimplification, and selective framing are used to mislead as it mocks the idea of clean water as a priority, ignoring the serious health concerns that prompted the government to consider these reforms, downplays the significance of regulatory standards by sarcastically framing them as if they are unnecessary, while the actual cost estimates are not addressed seriously, dismisses the efficiency gains oversimplifying the government's rationale for proposing larger entities to handle water services and sarcastically dismisses the complexity of reviewing all three waters, suggesting that it is unnecessarily complicated. 

Pre-trained Large Language Models (LLM) and deep neural networks have been acknowledged as efficient and effective techniques to address the framing classification and misinformation detection problem since they can learn from unstructured data and identify complex patterns that are difficult to detect using traditional methods \cite{islam2020deep}. 

Our hypothesis is that news or articles on the same topic can be converted into misinformation when given different frames, and in this paper, we use the frame of a narrative and the frame elements as key considerations in the process of identifying misinformation.

Our contributions of this research work include:

\begin{itemize}
    \item We formally define misinformation that is portrayed from the facts and formulate the misinformation detection problem in the context of Generative AI. 
    
    \item We propose a novel model called Framed Element-based Model (FEM), which can effectively identify misinformation stemming from portrayed facts under different framing. To the best of our knowledge, this is the first full research work, tackling the framing-based misinformation detection problem.
    
    
    \item We are the first to investigate how framing elements affect misinformation detection, treating each element as a separate feature for the language model to process. Our research systematically examines these elements, offering important insights into the subtle ways information can be skewed using framing. This also enhances the accuracy and effectiveness of detecting misinformation.
    
\end{itemize}

The rest of this work is organized as follows. In Section \ref{sec:relatedwork} we discuss related works by examining misinformation detection and framing theory. In Section \ref{sec:preliminary}, we give the formal definitions and formulate the problem. Our proposed FEM model is then explained in Section \ref{sec:model}. Experimental setups and datasets are introduced in Section \ref{sec:dataset}. In Section \ref{sec:experiments}, four experiments are conducted to evaluate the proposed model, analyze the parameters of the four framing elements, 
and introduce a case study which provides tangible illustrations of the model's effectiveness. Lastly, in Section \ref{sec:conclusion}, we conclude the paper and give recommendations for future work.

\section{Related Work} \label{sec:relatedwork}

\subsection{Traditional Misinformation Detection}

With the rise of social media, the ease with which information can be distributed and consumed has increased, allowing misinformation also to increase \cite{islam2020deep}. Traditional rule-based misinformation detection for fact-checking and fake news focused on detecting misinformation by focusing on who provided the information or what the content of the information was. Manual fact-checking relied on the author's reputation and/or the source to determine the veracity of the information \cite{guo2022survey}. Similarly, to detect fake news on social media, the social contexts, such as explicit and implicit features of user’s profiles, are evaluated to determine the credibility of the information \cite{shu2018understanding}. In addition to social contexts, fake news detection focuses on the content of the text by extracting linguistic features in order to detect sensational headlines that are frequent in fake news \cite{shu2018understanding}. Moreover, identifying negation keywords, such as `no,' `not,' or `never,' played a significant role in enhancing the classification of rumors \cite{kwon2013prominent}. Traditional rule-based approaches relied on information specific to the topic to correctly identify misinformation, therefore, these approaches experienced limitations when detecting misinformation about a new topic \cite{vlachos2014fact}. These shortcomings were addressed with the introduction of semi-supervised and unsupervised methods \cite{oshikawa2018survey}.

\subsection{Deep Learning Based Misinformation Detection}

Many researchers have explored the use of deep learning techniques to automate misinformation detection, such as tensor and transformer-based models and convolutional and recurrent neural networks \cite{islam2020deep,abdali2020tensor,nasir2021fake,pelrine2021surprising, pillai2023misinformation}. Latent patterns and spatial context were extracted from tensor-based models to construct k-nearest-neighbour graphs and belief propagation for semi-supervised misinformation detection \cite{abdali2020tensor}. A hybrid of convolutional neural networks (CNN)  and recurrent neural networks (RNN) leverages the strengths of CNN in extracting local features and of RNN in capturing long-term dependencies to detect fake news \cite{nasir2021fake}. Another RNN model found that combining sentiment, emotional, irony and hate analysis with bagging, boosting, stacking and voting means, produced a higher accuracy than without the various analyses \cite{pillai2023misinformation}. An evaluation of transformer-based models Large Language Models, namely, BERT variants to be used as baselines for misinformation detection, can achieve comparable or better performance than more complex state-of-the-art methods \cite{pelrine2021surprising}. More recently, a transformer-based model, MisRoB\AE RTa, utilized RoBERTa and BART to outperform single transformer misinformation detection models \cite{truica2022misrobaerta}. Finally, a hybrid deep learning model integrating features-based models and universal sentence encoding revealed promising results on the PHEME dataset \cite{alzahrani2023hybrid}.

While these techniques are able to accurately detect misinformation without considering the narrative or frame, their challenge lies in dealing with misinformation stemming from factual events that are skewed to convey a different implication. Furthermore, they also face difficulties handling lengthy news articles that potentially contain both truthful and misleading information.

\subsection{Framing Theory}
The frame of a piece of text can increase the salience of specific parts of information, i.e., to make information more meaningful, noticeable, or memorable \cite{entman1993framing}. An example by Entman showed that a frame can influence how a large portion of readers notice, understand, remember, evaluate, or act upon information presented to them \cite{entman1993framing}. According to Entman, the problem definition, causal interpretation, moral evaluation, and treatment recommendation are the four identifiable elements of a frame \cite{entman1993framing}. Multiple methods have been developed to detect frames using different approaches. Liu et al. detected frames from news based on the article headlines by fine-tuning a Large Language Model, i.e., BERT \cite{liu2019detecting}. Alternatively, Walter and Ophir leveraged computational tools to develop a novel method, the Analysis of Topic Model Networks, for the inductive identification and categorization of frames \cite{walter2019news}. Although both misinformation, frame, and framing element detection are possible, the impact of frames on misinformation detection requires further research. Our proposed FEM explores this impact by incorporating the framing theory presented by Entman to solve the earlier challenge of detecting misinformation stemming from accurate facts that are skewed to be misleading potentially \cite{entman1993framing}. Additionally, FEM discerns the respective contributions of the four framing elements to the overall accuracy of misinformation detection.

\section{Preliminary}\label{sec:preliminary}

In this section, we give formal definitions, and the problem of detecting misinformation portrayed from the facts is also formulated.

\subsection{Formal Definition}

\noindent
\textbf{Definition 1: Narrative} generally refers to a way of sharing stories or information, whether it is spoken, written, or shared online. In the current context, the narrative indicates the news stories and articles that are being disseminated online. Let $\mathcal{N} = \{n_1, n_2, \ldots, n_n\}$ denote the set of narratives, a narrative can be information or misinformation in online social networks, where $n_i$ represents a single narrative. \\

\noindent
\textbf{Definition 2: Information} refers to the presentation of facts in a way that aims to convey these facts accurately. The narrative of the information is constructed to reflect its true nature and implications without distorting or omitting key elements. Let $\mathcal{I} \in \mathcal{A}$ refer to the information set:

\begin{equation}
\mathcal{I} = \{ (fa, n) \mid fa \in \mathcal{FA} \wedge n \in \mathcal{N} \}
\end{equation}

\noindent
where $fa \in \mathcal{FA}$ refers to a specific fact in a fact set, $n \in \mathcal{N}$ refers to a narrative of a narrative set for information.\\

\noindent
\textbf{Definition 3: Misinformation}, converse to information, involves using the same facts but framing them within a narrative that is designed to mislead, deceive, or manipulate the audiences. The key aspect of misinformation in this work is not the distorted facts, but how they are presented in a misleading narrative. Let $\mathcal{M} \in \mathcal{A}$ represent a set of misinformation which is composed of the content and the specific narrative:

\begin{equation}
\mathcal{M} = \{ (fa, n) \mid fa \in \mathcal{FA} \wedge n \in \mathcal{N} \}
\end{equation}

\noindent
where $fa \in \mathcal{FA}$ refers to the same fact in a fact set as information, $n \in \mathcal{N}$ refers to the narrative of a narrative set for misinformation.\\

\noindent
\textbf{Definition 4: Frame} suggests how information is structured and presented in a story, including the perspective from which it is told. It suggests how information is presented in a narrative - the story that communicates the facts in a meaningful way - can influence an individual's perception and interpretation of that information and is recognized as an important concept in the communication and social science fields. Mathematically, $f$ represents a frame, and the set of frames is $\mathcal{FR} = \{fr_1, fr_2, \ldots, fr_m\}$. The relationship between a frame and a narrative of one article can be represented by $\mathcal{R}: \mathcal{N} \rightarrow \mathcal{FR}$, where $R(n_i) = fr_i$, and $\mathcal{R}$ represents the element extractor. \\

\noindent
\textbf{Definition 5: Frame Elements} are the specific components used to construct a frame in articles. A frame is generally composed of four elements, and they constitute how information should be displayed in front of the readers and how the readers would perceive the content. Each article has four elements: “problem definition”, “causal interpretation”, “moral evaluation”, and “treatment recommendations”. Let $e$ represent one of the elements of a frame in an article and $\mathcal{E}_i = \{\text{e}_{1}, \text{e}_{2}, \text{e}_{3}, \text{e}_{4}\}$ represents the element set of the article $a_i$ where $e_1$ represents the “problem definition”, $e_2$ represents the “causal interpretation”, $e_3$ represents the “moral evaluation”, and $e_4$ represents the “treatment recommendations”.

\subsection{Problem Formulation}
The Misinformation Detection problem is defined as the process of classifying articles to identify misinformation stemming from portrayed facts under different narratives thus misleading the audiences. To achieve that, we adopt the Frame Element-based Model (FEM) incorporating the elements of framing theory extracted from the articles. The FEM is trained to understand the semantics and narratives of articles. Having a set of articles $ \mathcal{A}=\{a_i, a_2, \ldots, a_n\}$, given an article $a_i \in A$, the model first extracts the frame elements $E_i=\{e_1, e_2, e_3, e_4\}$ of the article, and then encode them to get the hidden state $h_i$ of it which is later used to calculate the probability to predict if the article is misinformation or not.

\begin{equation}
P(h_i) = softmax(w\cdot h_i + b),
\end{equation}

\noindent
where $P(h_i)$ is the probability that an article $a_i$ contains misinformation, and $h_i$ represents the last hidden state of the given article $a_i$ or corresponding element set $E_i$, .

The object of the last step of predicting is defined as minimizing the loss function $\mathcal{L}$:

\begin{equation}
\mathbf{w}^*, b^* = \arg\min_{\mathbf{w},b} \mathcal{L}(\mathbf{w}, b) + \lambda \|\mathbf{w}\|^2,
\end{equation}

\noindent
where $w^*$ and $b^*$ are the target optimal weights, and the loss function $\mathcal{L}$  which is the cross entropy loss function is defined as:

\begin{equation}
\mathcal{L} = -\frac{1}{N} \sum_{i=1}^{N} \left[y_i \log P(h_i) + (1 - y_i) \log(1 - P(h_i))\right],
\end{equation}

\noindent
where $N$ is the number of samples, and $y_i$ is the actual label of the article $a_i$.

\section{Frame Element-based Misinformation Detection Model}\label{sec:model}

\begin{figure}[htbp]
    \centerline{\includegraphics[width=0.5\linewidth]{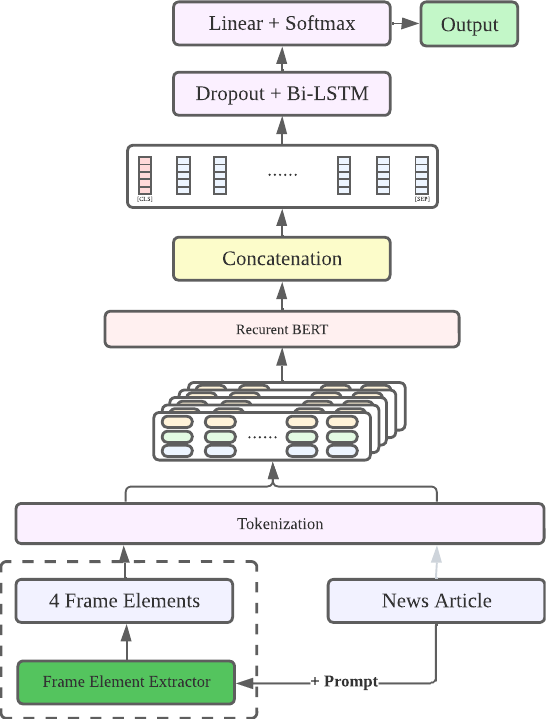}}
    \caption{The Architecture of the Frame Element-based Model}
    \label{fig:overall_architecture}
\end{figure}

In this section, the proposed Frame Element-based Model (FEM) for Misinformation detection is introduced in the context of news articles. Figure~\ref{fig:overall_architecture} demonstrates the overall framework of our proposed model, and in Algorithm~\ref{alg:FEM}, we showcase the steps of the whole process.

Initially, the Frame Element Extractor is utilized to process the news article to extract four framing elements: Problem Definition, Causal Interpretation, Moral Evaluation, and Treatment Recommendation. These elements represent the core of how the information is framed. The extracted framing elements, along with the corresponding news article, are then tokenized performing as the fundamental preprocessing step in NLP.

To capture the subtle contextual nuances of each element, following the tokenization, we independently encode each element and the corresponding news article (Lines 6 to 14 in Algorithm~\ref{alg:FEM}). This nuanced understanding of different elements is vital as each frame carries different weights and implications for the overall narrative of this news article. The separate encoding also allows us to quantify the impact of each element to reveal the most influential aspects of how the article is framed thus increasing the chance of identifying misinformation. 

Independently encoding each element and article is a strategic choice that can enhance the model's analytical precision, allowing for reliable misinformation detection. Line 6 starts the recurrent process. An empty tensor is created in advance and used to concatenate each embedding from each loop. In the recurrent process, we first encode the article which performs as the main body of the input, and then each element is encoded.

The embeddings of each element $embE$ and the news article $embA$ are then concatenated to form a dense vector $emb$ as the representation of the whole input followed by a dropout layer to prevent overfitting.

\begin{equation}
emb_t=concat(emb_{t-1},h_t), 
\end{equation}

\noindent
where $emb_t$ represents the concatenated embeddings of the current time step, and $h_t$ represents the embeddings of an element $embE$ or the article $embA$.

The concatenated embeddings $emb$ from the previous layer are then fed to a Bi-LSTM layer. The Bi-LSTM layer is applied for the purpose of capturing the holistic context after all elements and article embeddings are concatenated. This allows the model to understand how different elements of the article relate to each other.

\begin{equation}
h_i=Relu(BiLSTM(emb)),
\end{equation}

\noindent
where $h_i$ is the representation of the input after being processed by a Bi-LSTM layer and a Relu activation function.

A linear layer including a Dropout is applied to map the high-dimensional output representation $h_i$ from the Bi-LSTM layer to the target space. The softmax function is used to obtain the probability distribution over the potential classes, which finalizes the prediction process to identify the misinformation.

\begin{equation}
predicts=softmax(Dropout(h_{i})W^{T}+b),
\end{equation}

\noindent
where $predicts$ is the probability distribution of the class labels, $W^{T}$ is the learnable weight matrix, $b$ refers to the bias.

\begin{algorithm}
	\caption{Frame Element-based Misinformation detection Algorithm}\label{alg:FEM}
		Input: $D=(a,\ E)$  
		\\
		Output: 0 (misinformation) or 1 (information)
		\begin{algorithmic}[1]
			\State $Information \gets Collect(sources)$
                \State $Misinformation \gets ChatGPT(information,\ prompt1)$
                \State $FrameElements \gets ChatGPT(articles,\ prompt2)$
                \State Create BERT, BiLSTM, FC Layer as classifier, Dropout, Relu
                \State $emb: = \{\}$
			\For {$a_i, E_i \in D$}
			    \State $embA: = BERT(a_i)$
                    \State $emb: = concat(emb,\ embA)$
    			\For {$e_j \in E_i$}
                        \State $embE: = BERT(e_j)$
                        \State $emb: = concat(emb,\ embE)$
    			\EndFor
                    \State $emb: = Dropout(emb)$
			\EndFor

                \State $outputs: = BiLSTM(emb)$
                \State $h: = Relu(outputs)$
                \State $logits: = classifier(Dropout(h))$
			\State $predicts: = softmax(logits)$
		\end{algorithmic}
	\end{algorithm}

\section{Experiment Setups}\label{sec:dataset}
\subsection{Model Setup}

To ensure an efficient training process, we conduct our experiments on the Paperspace \footnote{https://www.paperspace.com/} platform utilizing the following tailored computational and training settings to the unique demands of each dataset:
\begin{itemize}
  \item GPU Configuration: The model is trained over a span of 100 epochs utilizing NVIDIA's A6000 48GB GPU and 45 GB 8 CPU.
  \item Dropout: To mitigate the risk of overfitting, a dropout rate of \(0.3\) was applied during training. 
  \item Learning Rate: The training uses an initial learning rate of \(1 \times 10^{-5}\) and it is modulated following a cosine schedule with a warm-up phase. The warm-up steps vary in accordance with the specificities of each dataset.
  \item Batch Size: The batch size is determined based on the particular requirements and characteristics of each dataset.
  \item Frame Element Extractor: ChatGPT, as a powerful generative AI model, is used as the element extractor. Different extractors can be applied for the same purpose.
\end{itemize}

\subsection{Datasets}
In this section, we introduce 4 datasets used to evaluate our model. To assess the generalization capability of the model, we used three single-topic datasets which are the Three Waters Reform dataset, Covid-19 dataset, Nuclear Pollution dataset, and a mixed-topic dataset which is the Kaggle Fake News dataset. The statistics of these datasets are displayed in Table~\ref{tab:dataset_statistics}.

\begin{itemize}
    \item \textbf{Three Waters Reform} dataset is collected from The Knowledge Basket\footnote{https://www.knowledge-basket.co.nz/}. We only capture the news focus on the ``Three Waters Reform" in New Zealand, a topic of substantial political discourse and interest spanning from 2017 to June 2023. This dataset accumulates a total of 1,841 articles. Following the application of our labeling process yields 3,262 articles labeled in concordance with their identified frames and frame elements.

    \item \textbf{Covid-19} is collected using Newsapi \footnote{https://newsapi.org/} which is an API service that allows developers to retrieve news articles from various sources on a worldwide scale. We use ``Covid-19" as keywords to retrieve news articles in the period from 01/12/2019 to 20/08/2023. These articles reflect the in-time attitude to the Covid-19 pandemic. This dataset includes 13,386 articles after the pre-processing.
    
    \item \textbf{Nuclear Pollution} dataset is collected using the Newsapi as well and with the keywords ``nuclear pollution" over the last 5 years. This dataset provides a comprehensive view of the discourse surrounding nuclear pollution offering a diverse range of perspectives and information. After the data pre-processing, there are 2,431 articles with an average token length of 482.
    
    \item \textbf{Kaggle Fake News Dataset} \footnote{https://www.kaggle.com/datasets/stevenpeutz/misinformation-fake-news-text-dataset-79k} contains news articles from multiple sources such as Reuters and so on. For the purpose of our study, we confine our selection to the ``TRUE" set and randomly select 3k articles. To augment the dataset to fulfill the research objectives, we produce another set of data by varying the frame of existing news, ultimately resulting in 5,915 labeled samples following the implementation of our augmentation process.
\end{itemize}

\renewcommand{\arraystretch}{1.5} 
\begin{table}[!ht]
\caption{The statistics of the datasets after pre-processing.}
\label{tab:dataset_statistics}
\centering
\begin{tabular}{l|cc}
\hline
Dataset & articles & average length         \\ \hline
The Three Waters      & 3,262  & 823          \\ \hline
Covid-19              & 13,386 & 537          \\ \hline
Nuclear Pollution     & 2,431  & 482           \\ \hline
Mixed-topic           & 5,915  & 469           \\ \hline
\end{tabular}
\end{table}

\subsection{Data Pre-processing}

Our proposed methodology commences with the collection of datasets comprised of news articles from reliable sources. These articles constitute our ground truth, representing information opposite to misinformation. Accordingly, we synthesize misinformation based on the framing theory by altering the frames of our collected news articles. This process augments our datasets in a generative method at the document level and unfolds in three structured phases \cite{bayer2022survey}.

\begin{itemize}
    \item \textbf{Frame Identification and Element Extraction}: Utilizing the capabilities of ChatGPT, we first process the collected news articles to identify their frames and extract four elements of framing theory. These extracted framing elements reflect the news articles' original and unaltered state. They are annotated with the label ``1", signifying their category as information. The frames we harness in this work are selected and proved by domain experts in communication.

    \item \textbf{Frame Alteration}: The second stage involves the alteration of the frame, utilizing ChatGPT to manipulate the article narrative while maintaining the original factual information. This step simulates the process of creating misinformation through narrative manipulation, a common way that preserves factual information but skews the frame to mislead readers. 20\% of the altered narratives are verified by the domain experts.

    \item \textbf{Element Extraction}: In the final step, we process the narrative-manipulated articles through ChatGPT to extract the corresponding four elements of framing theory, labeled ``0" along with the manipulated articles, signifying their category as misinformation. This eventually establishes the basis for comparison with the information.

\end{itemize}

This pre-processing procedure is designed to construct binary-category datasets that are comprised of information and misinformation with elements of framing theory, thereby enabling the nuanced training of our model. Through this process, we not only aim to create datasets that serve as the foundation of misinformation detection but also enhance the understanding of how narrative (framing theory in this work) can be utilized to generate misinformation.

\subsection{Evaluation Metrics and Baselines}
To evaluate the performance of our proposed model (FEM), we utilize the Confusion Matrix as our primary evaluation measurement. The Confusion Matrix provides a comprehensive visualization of the performance by categorizing predictions into four different classifications \cite{shu2017fake}: 
\begin{itemize}
    \item True Positives (TP): when predicted misinformation is actually labeled as misinformation; 
    \item True Negatives (TN): when predicted information is actually labeled as information;
    \item False Positives (FP): when predicted information is actually labeled as misinformation; 
    \item False Negatives (FN): when predicted misinformation is actually labeled as information.
\end{itemize}

As our baselines, we perform the fine-tuning of several highly utilized pre-trained transformer-based language models followed by a feed-forward layer as a classifier for misinformation detection. These baselines include: 

    \begin{itemize}
        \item \textbf{BERT} \cite{devlin2018bert} is a groundbreaking transformer-based model in the field of natural language processing (NLP). It is known for its deep bidirectional training, meaning it considers the context from both the left and right sides in all layers. This leads to a more nuanced understanding of language context and semantics. BERT has been highly influential in improving the performance of a wide range of NLP tasks.

        \item \textbf{RoBERTa} \cite{liu2019roberta} is built upon BERT by modifying key hyperparameters, training with more data, and longer training times. These changes help RoBERTa outperform BERT on several benchmark NLP tasks. It is known for its improved robustness and efficiency.

        \item \textbf{ALBERT} \cite{lan2019albert} is a version of BERT optimized for lower memory consumption and increased speed. It introduces two major modifications: factorized embedding parameterization and cross-layer parameter sharing. These changes reduce the model's size without significantly affecting its performance, making it more scalable and efficient.

        \item \textbf{XLNet} \cite{yang2019xlnet} is an extension of the Transformer model. Instead of the standard transformer, XLNet uses transformer-XL \cite{dai2019transformer}. It combines the best of both autoregressive (AR) and autoencoding (AE) models. Unlike BERT, XLNet learns to predict a word at a position in a sequence considering all permutations of the sequence.

        \item \textbf{LongFormer} \cite{Beltagy2020Longformer} is designed to handle longer texts. It is an extension of the standard transformer-based model, like BERT, but optimized for lengthy documents. Its key innovation is the introduction of an attention mechanism that scales linearly with sequence length.

    \end{itemize}

\section{Experiments and Analysis}\label{sec:experiments}
\subsection{Experiment 1: Model Evaluation - against baselines}

In this experiment, we compare the performance of our model with other baseline models on four datasets introduced in Section \ref{sec:dataset}. The results are displayed in Table~\ref{tab:exp1_three_water} ,\ref{tab:exp1_covid_19}, \ref{tab:exp1_nuclear} and \ref{tab:exp1_mixed} respectively. The results on each dataset consistently show that our model (FEM) incorporating frame elements with the original news article significantly outperforms other models with only articles, presenting the importance of frame elements.

From these results, we can observe that frame elements play a crucial role in understanding and interpreting information. We also demonstrate the results of our model with only frame elements and only texts as input respectively. Compared to the baselines, the results with only frame elements as input are also beyond them.

By analyzing the performance of our model with only frame elements compared to the other baselines, we can observe the significance of these elements. Frame elements contribute to the deeper semantic understanding of the content by narrowing it down to the core theme reducing the distracting noise. This enables the model to grasp not just the explicit meaning but also the implicit intentions and nuances thus increasing the probability of precisely detecting misinformation.

Experiment results in Table~\ref{tab:exp1_mixed} on the Mixed-topic dataset demonstrate that frame elements can also provide a more general representation of information, making the model more adaptable and robust to variations of information.

\renewcommand{\arraystretch}{1.5} 
\begin{table}[h!]
\caption{Results on the Three Waters Dataset.}

\centering
\begin{tabular}{c|cccc}
\hline
Models    & Accuracy & Precision & Recall & F1\_score \\ \hline
BERT      & 0.8469   & 0.8188    & 0.8127 & 0.8157         \\ \hline
RoBERTa   & 0.8622   & 0.8784    & 0.7915 & 0.8327         \\ \hline
ALBERT    & 0.8086   & 0.7651    & 0.8057 & 0.7849         \\ \hline
XLNet     & 0.8545   & 0.8113    & 0.8657 & 0.8376         \\ \hline
LongFormer & 0.8591   & 0.8283    & 0.8516 & 0.8398         \\ \hline
FEM (text+frames) & \textbf{0.9862}   & \textbf{0.9695}    & \textbf{0.9734} & \textbf{0.9715}   \\ \hline
FEM (only text) & 0.8652   & 0.8316    & 0.8638 & 0.8474         \\ \hline
FEM (only frames) & 0.9278   & 0.9355    & 0.9605 & 0.9478         \\ \hline
\end{tabular}
\label{tab:exp1_three_water}
\end{table}

\renewcommand{\arraystretch}{1.5} 
\begin{table}[h!]
\caption{Results on the Covid-19 Dataset.}
\centering
\begin{tabular}{c|cccc}
\hline
Models    & Accuracy & Precision & Recall & F1\_score \\ \hline
BERT      & 0.8372   & 0.8052    & 0.8074 & 0.8063         \\ \hline
RoBERTa   & 0.8547   & 0.8539    & 0.7867 & 0.8190         \\ \hline
ALBERT    & 0.8104   & 0.7783    & 0.8163 & 0.7968         \\ \hline
XLNet     & 0.8429   & 0.8207    & 0.8629 & 0.8412         \\ \hline
LongFormer & 0.8546   & 0.8617    & 0.8694 & 0.8655         \\ \hline
FEM (text+frames) & \textbf{0.9783}   & \textbf{0.9583}    & \textbf{0.9708} & \textbf{0.9645}         \\ \hline
FEM (only text) & 0.8865   & 0.8737    & 0.8826 & 0.8781         \\ \hline
FEM (only frames) & 0.9132   & 0.9195    & 0.9361 & 0.9277         \\ \hline
\end{tabular}
\label{tab:exp1_covid_19}
\end{table}

\renewcommand{\arraystretch}{1.5} 
\begin{table}[h!]
\caption{Results on the Nuclear Pollution Dataset.}
\centering
\begin{tabular}{c|cccc}
\hline
Models    & Accuracy & Precision & Recall & F1\_score \\ \hline
BERT      & 0.8035   & 0.7921    & 0.80167 & 0.7969         \\ \hline
RoBERTa   & 0.8167   & 0.8234    & 0.7826 & 0.8025         \\ \hline
ALBERT    & 0.8051   & 0.7568    & 0.7864 & 0.7713         \\ \hline
XLNet     & 0.8268   & 0.8035    & 0.8284 & 0.8158         \\ \hline
LongFormer & 0.8462   & 0.8254    & 0.8316 & 0.8285         \\ \hline
FEM (text+frames) & \textbf{0.9538}   & \textbf{0.9429}    & \textbf{0.9531} & \textbf{0.9480}         \\ \hline
FEM (only text) & 0.8491   & 0.8365    & 0.8537 & 0.8450         \\ \hline
FEM (only frames) & 0.9035   & 0.9216    & 0.9268 & 0.9242         \\ \hline
\end{tabular}
\label{tab:exp1_nuclear}
\end{table}

\renewcommand{\arraystretch}{1.5} 
\begin{table}[h!]
\caption{Results on the Mixed-topic Dataset.}
\centering
\begin{tabular}{c|cccc}
\hline
Models    & Accuracy & Precision & Recall & F1\_score \\ \hline
BERT      & 0.8354   & 0.8127    & 0.8165 & 0.8146         \\ \hline
RoBERTa   & 0.8497   & 0.8503    & 0.7902 & 0.8191         \\ \hline
ALBERT    & 0.8126   & 0.7816    & 0.8257 & 0.8030         \\ \hline
XLNet     & 0.8528   & 0.8320    & 0.8783 & 0.8545         \\ \hline
LongFormer & 0.8736   & 0.8542    & 0.8867 & 0.8701         \\ \hline
FEM (text+frames) & \textbf{0.9696}   & \textbf{0.9582}    & \textbf{0.9683} & \textbf{0.9632}         \\ \hline
FEM (only text) & 0.8823   & 0.8574    & 0.8929 & 0.8748         \\ \hline
FEM (only frames) & 0.9158   & 0.9207    & 0.9319 & 0.9263         \\ \hline
\end{tabular}
\label{tab:exp1_mixed}
\end{table}

\subsection{Experiment 2: Parameter Analysis}

In this experiment, we conduct a comparative analysis to explore the contribution of each element to misinformation detection. The experimental framework analyzes the composite efficacy of the model equipped with the four elements, i.e., Problem Definition, Causal Interpretation, Moral Evaluation and Treatment Recommendation. Within the area of misinformation detection with frame elements, exploring the individual contribution of distinct frame elements is vital to help us understand how frame elements influence the model's capability to grasp the veracity of information.

The model with all 4 elements serves as the benchmark for optimal performance, showing a high degree of accuracy, precision, recall and F1-score. This provides a holistic frame element-based analysis of information, thus enhancing the probability of identification of misinformation.

\begin{figure}[htbp]
    \centering
    \begin{subfigure}{.45\textwidth}
        \centering
        \includegraphics[width=\linewidth]{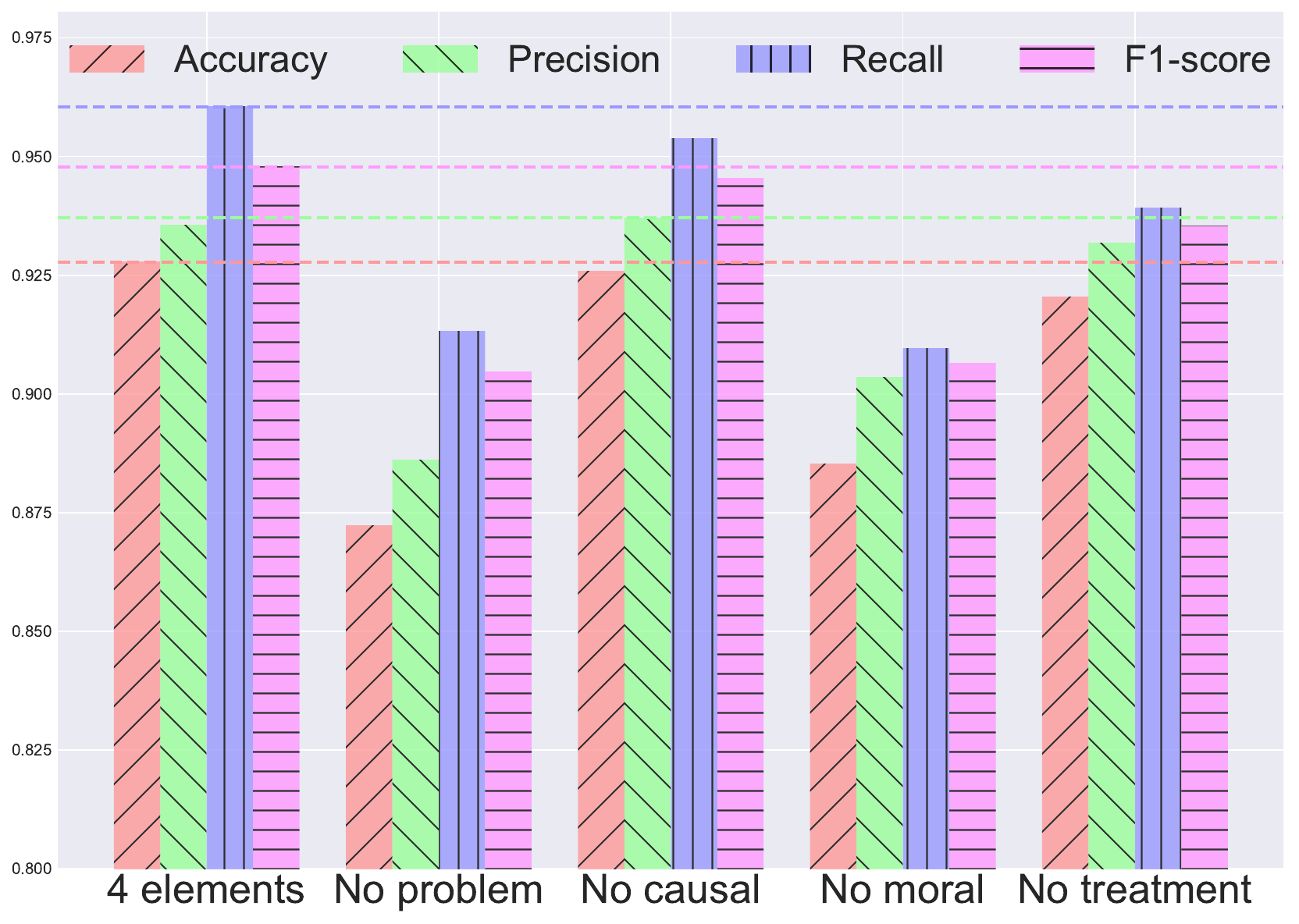}
        \caption{Three Waters Reform}
        \label{fig:ablation_four_elements_twr}
    \end{subfigure}
    \begin{subfigure}{.45\textwidth}
        \centering
        \includegraphics[width=\linewidth]{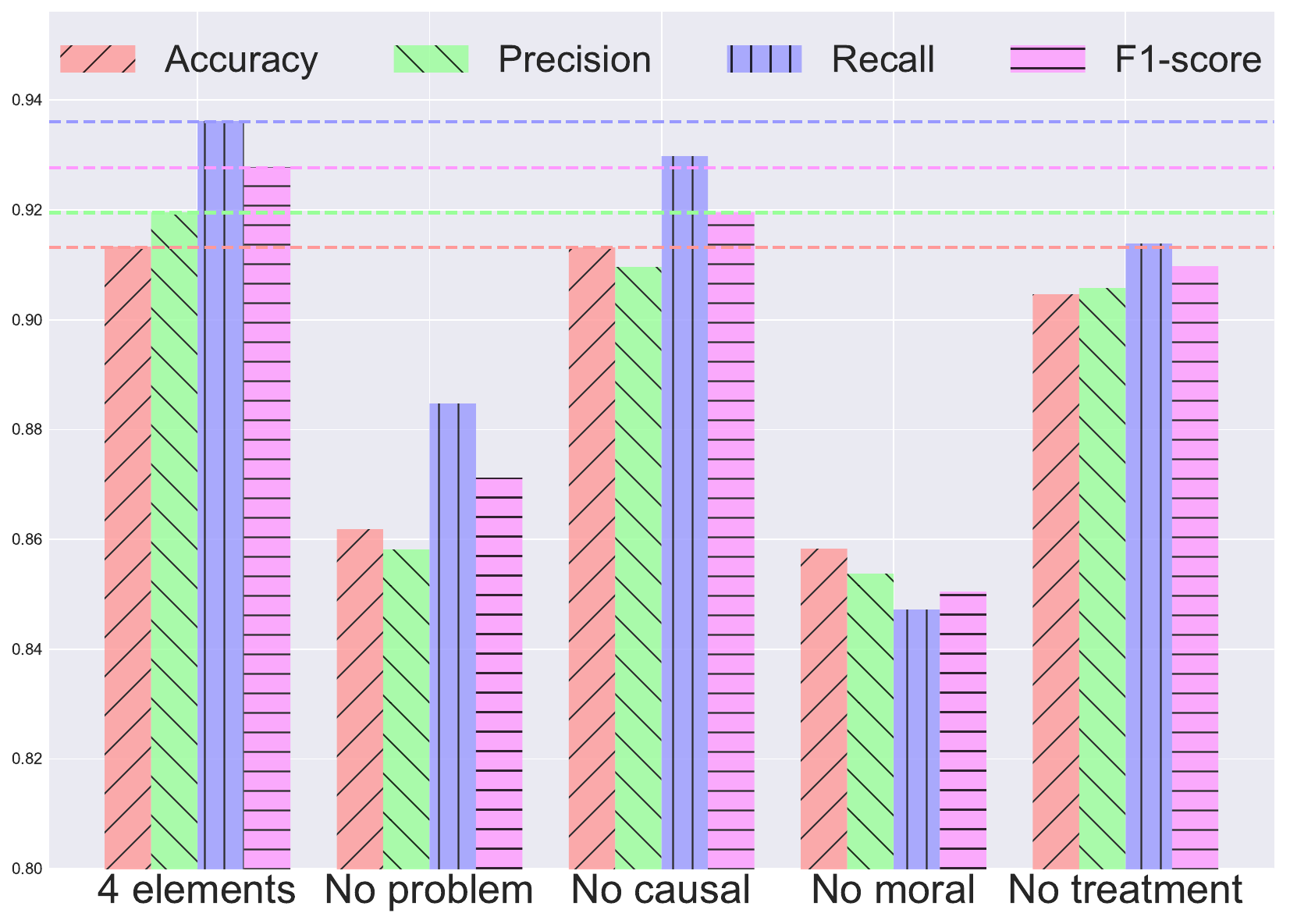}
        \caption{Covid-19}
        \label{fig:ablation_four_elements_covid}
    \end{subfigure}

    \begin{subfigure}{.45\textwidth}
        \centering
        \includegraphics[width=\linewidth]{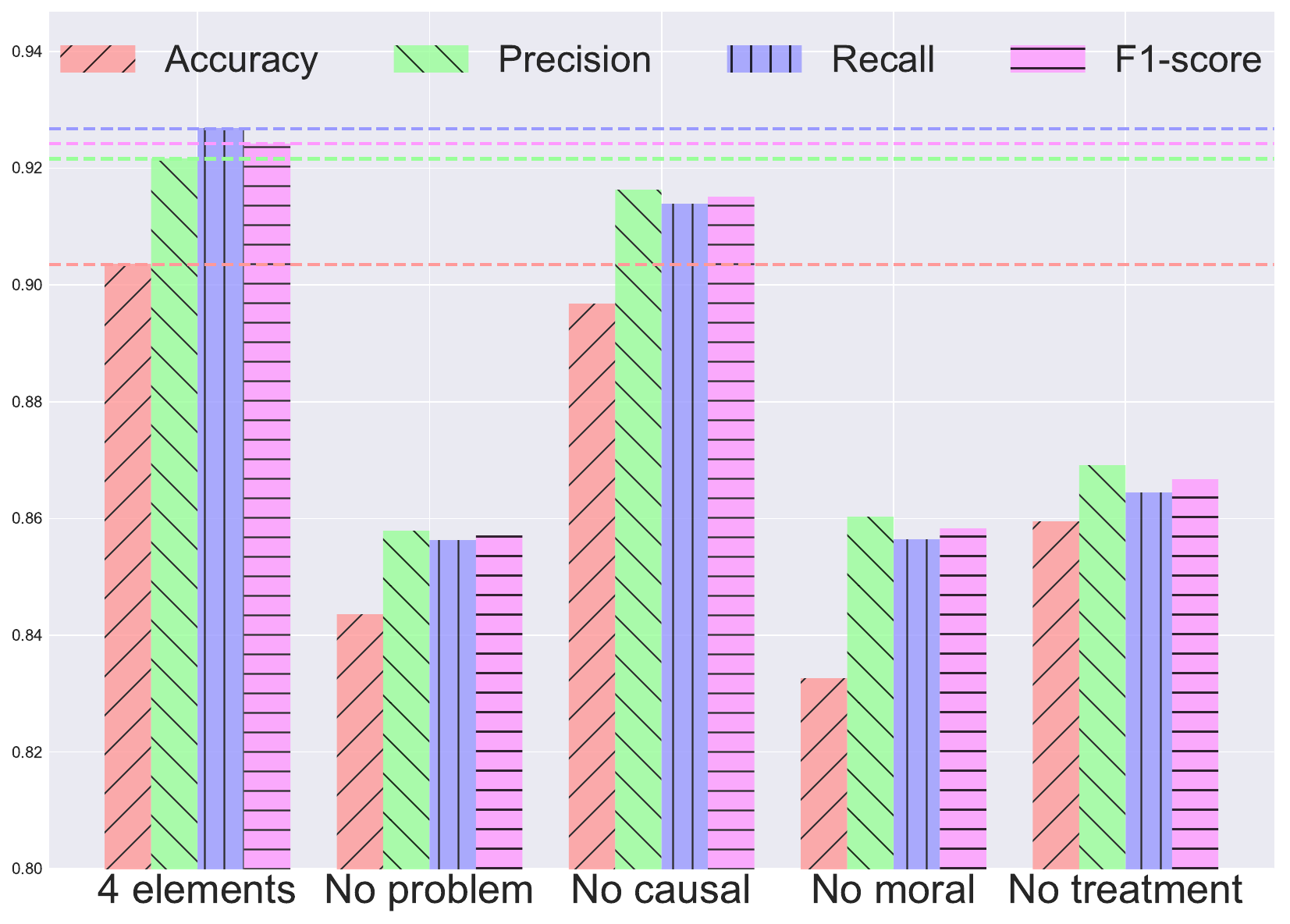}
        \caption{Nuclear Pollution}
        \label{fig:ablation_four_elements_nuclear}
    \end{subfigure}
    \begin{subfigure}{.45\textwidth}
        \centering
        \includegraphics[width=\linewidth]{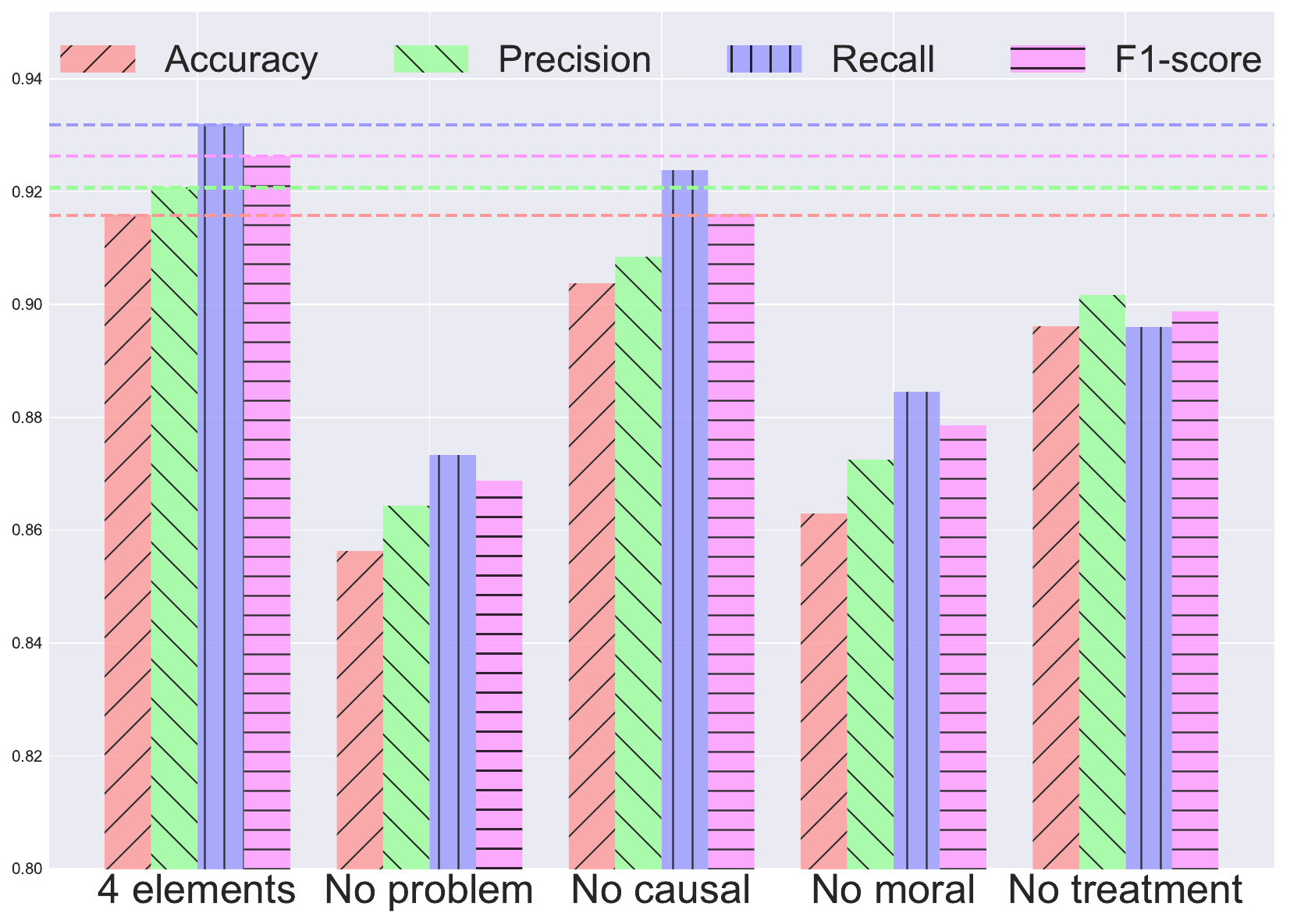}
        \caption{Mixed-topic}
        \label{fig:ablation_four_elements_mixed}
    \end{subfigure}
    
    \caption{Measure the performance of removing one of the elements on all four datasets.}
    \label{fig:ablation_four_elements}
\end{figure}

\begin{figure}[htbp]
    \centering
    \begin{subfigure}{.45\textwidth}
        \centering
        \includegraphics[width=\linewidth]{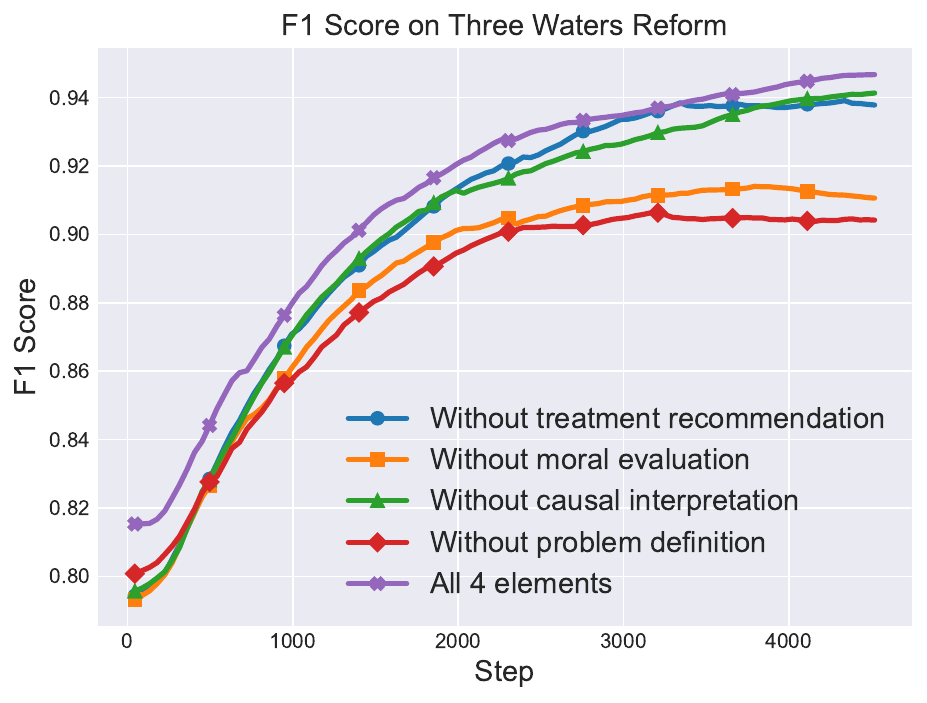}
        \caption{Three Waters Reform}
        \label{fig:f1_score_twr}
    \end{subfigure}
    \begin{subfigure}{.45\textwidth}
        \centering
        \includegraphics[width=\linewidth]{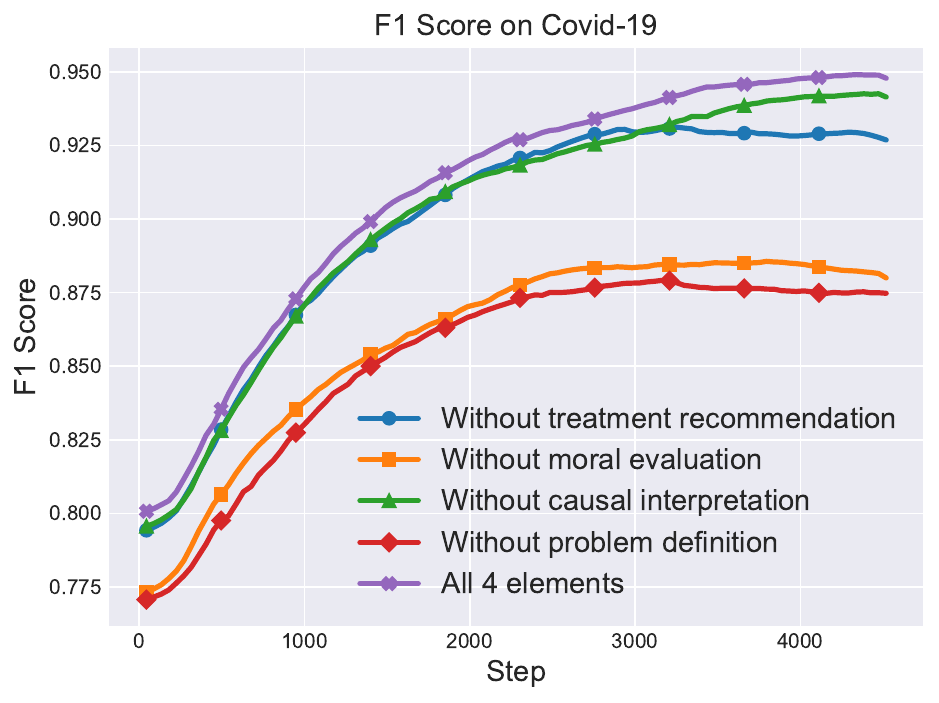}
        \caption{Covid-19}
        \label{fig:f1_score_covid}
    \end{subfigure}

    \begin{subfigure}{.45\textwidth}
        \centering
        \includegraphics[width=\linewidth]{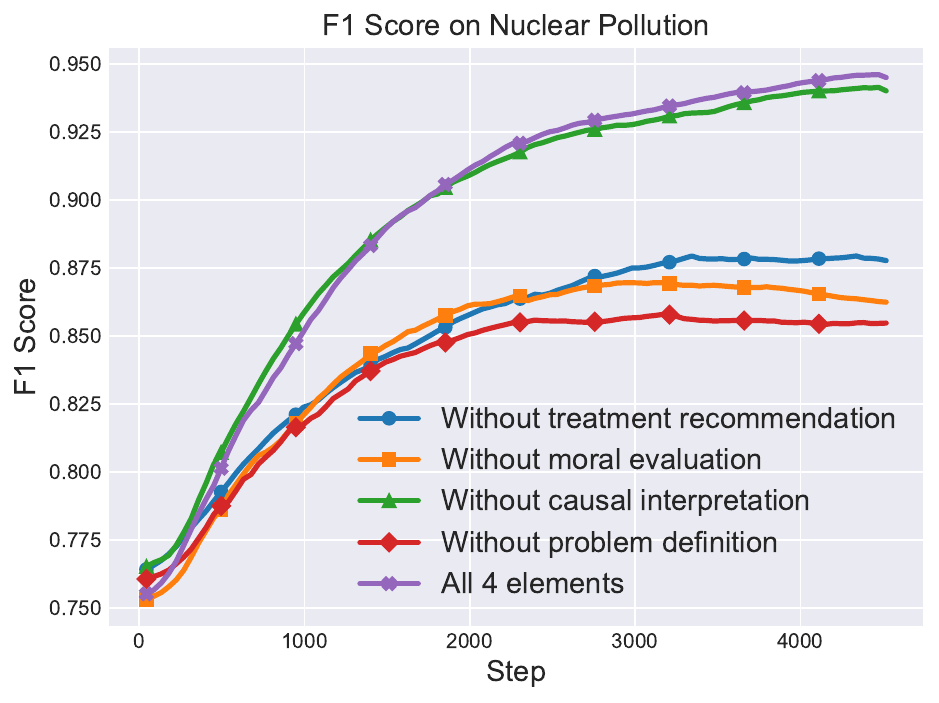}
        \caption{Nuclear Pollution}
        \label{fig:f1_score_nuclear}
    \end{subfigure}
    \begin{subfigure}{.45\textwidth}
        \centering
        \includegraphics[width=\linewidth]{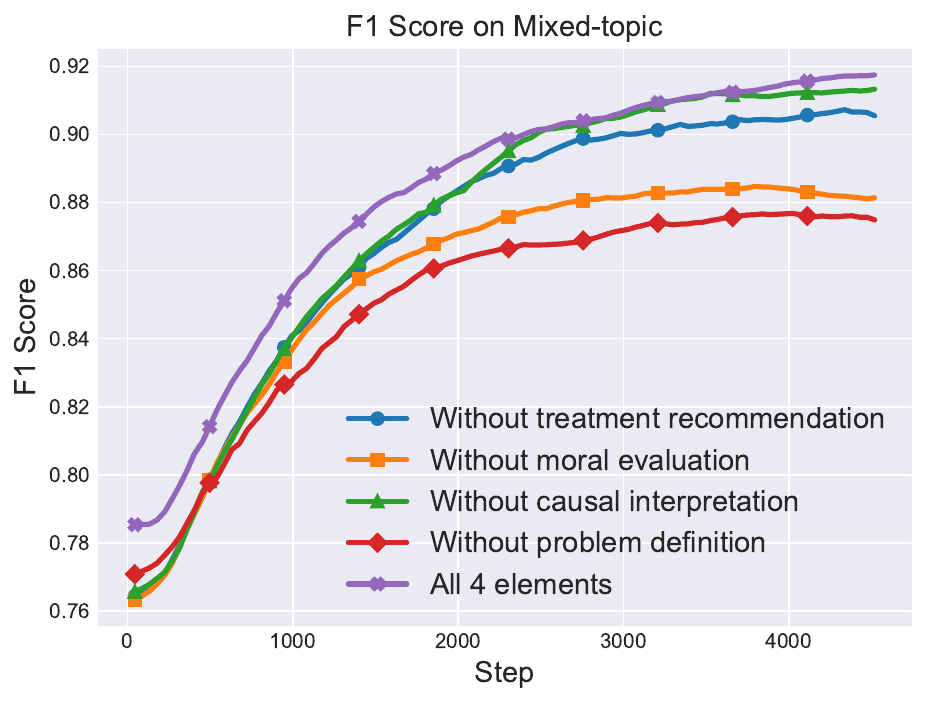}
        \caption{Mixed-topic}
        \label{fig:f1_score_mixed}
    \end{subfigure}
    
    \caption{The F1-scores during the training process on all four datasets.}
    \label{fig:F1_scores}
\end{figure}

Then, we remove each frame element from all four elements keeping the other three elements remaining. Figure~\ref{fig:ablation_four_elements} displays all performance metrics while Figure~\ref{fig:F1_scores} demonstrates the trend of F1-Scores during the training process.

We can observe from all these figures that when the element of Problem Definition is removed from the model, a pronounced decrement in all measurements is demonstrated. This suggests that the recognition of Problem Definitions is instrumental in the precise detection of misinformation, potentially due to its role in pinpointing the core theme within the narrative that may be manipulated. Without the incorporation of the element of Problem Definition, the capability of the model to differentiate between true and misleading content is significantly compromised.

Meanwhile, the absence of the frame of Moral Evaluation also results in a noticeable decline in all performance metrics. It appears to be an important factor in the framing of information indicating that it is often manipulated in misinformation in order to obtain emotional biases or ethical stances.

The model lacking the frame of Problem Definition or Moral Evaluation demonstrates a noticeable drop in Precision and Accuracy indicating a higher rate of false positives. This implies that while the model may still identify genuine instances of misinformation, it is also more likely to incorrectly classify accurate information as misinformation.

On the contrary, a lack of the frame of Causal Interpretation or Treatment Recommendation does not show a substantial decline in performance metrics compared to the benchmark. This observation implies that while they have a role in the misinformation detection process, however, their absence does not critically influence the capability of the model to identify misinformation.

One noticeable difference in the results demonstrated in Figure~\ref{fig:ablation_four_elements_nuclear} and Figure~\ref{fig:f1_score_nuclear} on the Nuclear Pollution dataset is the lack of the frame of Treatment Recommendation. Removing the Treatment Recommendation element also results in a lower performance across all metrics indicating that in the context of nuclear pollution, the treatment recommendation is likely to be a key indicator of the news articles. A lack of this element in this area could also allow misinformation proposing ineffective or misleading responses to harness the readers. The decrement in performance of missing the frame of treatment recommendation also implies that the contribution of each element can vary depending on the subject.

\subsection{Experiment 3: Similarity Comparison}

\begin{table}[tbp]
\caption{One single pair similarity and similarities removing one of the elements.}
\label{tab_single_similarity}
\centering
\scalebox{1.0}{%
\begin{tabular}{l|cc}
\hline
      Info vs Mis-info         & Similarity & F1-score \\ \hline
Article Similarity   & 0.86  & 0.8474              \\ \hline
Elements Similarity(all 4 elements) & \textbf{0.61} & \textbf{0.9478}  \\ \hline
Elements Similarity(without problem)   & 0.79  & 0.9046          \\ \hline
Elements Similarity(without causal)   & 0.62   & 0.9454         \\ \hline
Elements Similarity(without moral)   & 0.81    & 0.9065        \\ \hline
Elements Similarity(without treatment)  & 0.64  & 0.9354          \\ \hline
\end{tabular}}
\end{table}

\begin{table}[tbp]
\caption{Compare article average similarity with average similarity calculated using 4 elements.}
\label{tab_average_similarity}
\centering
\scalebox{1.0}{%
\begin{tabular}{l|cccc}
\hline
Info vs Mis-info & three water & covid & nuclear & mixed   \\ \hline
Article Similarity & 0.86 & 0.82 & 0.83 & 0.85\\ \hline
Elements Similarity(all 4 elements) & 0.58 & 0.62 & 0.59 & 0.61\\ \hline
Elements Similarity(without problem)   & 0.83 & 0.81 & 0.82 & 0.83           \\ \hline
Elements Similarity(without causal)   & 0.59 & 0.62 & 0.60 & 0.63            \\ \hline
Elements Similarity(without moral)   & 0.81 & 0.79 & 0.81 & 0.80            \\ \hline
Elements Similarity(without treatment)  & 0.60 & 0.64 & 0.78 & 0.63            \\ \hline
\end{tabular}}
\end{table}

In this experiment, we analyze the relationship between the similarity between information and misinformation under different conditions relating to the presence or absence of specific frame elements within our model. This experiment represents how closely misinformation mirrors authentic information in terms of framing. The cosine function is used to calculate their similarities:

 \begin{equation}
\label{eq:similarity}
sim(h_i,h_j) = \frac{h_i \cdot h_j}{\parallel h_i\parallel \parallel h_j \parallel},
\end{equation}

\noindent
where $h_i$ and $h_j$ represent the final hidden states of two articles or elements from two articles.

Table~\ref{tab_single_similarity} shows the similarities of one randomly selected article from the Three Waters Reform dataset and the overall performance (F1-score) of the model on this dataset. The similarity between the information article and the misinformation article is used as the benchmark for further analysis. The similarity of 0.86 shows that, without any specialized modification, misinformation is quite successful at resembling a genuine news article.

From Table~\ref{tab_single_similarity} we can observe that the similarity between the information article and misinformation article is the highest at 0.85, however, the F1 score of the model with only text on this dataset is 0.8474 which is lower than for other conditions especially when utilizing only all 4 elements which holds the lowest similarity 0.61 and highest F1-score 0.9478. 

This experiment indicates the alignment of similarity with the model performance as an inverse relationship. The lower the similarity, the higher the performance is in detecting misinformation. This underscores the importance of elements of framing theory in the detection process and points out their potential impact on improving detection accuracy.

We also calculate the average similarity scores between information and misinformation across four distinct datasets under different conditions. Results are displayed in Table~\ref{tab_average_similarity}. The pattern of similarity scores is relatively consistent across different topics, indicating that the manipulation of framing in misinformation follows a similar pattern. However, on the nuclear dataset, when omitting the frame of treatment recommendation, the similarity still remains at a high level, indicating the importance of treatment recommendation in this dataset. 

Overall, by comparing the average similarities across all datasets, we can observe the alignment of similarities with omitting distinct elements except the impact when omitting the frame of treatment recommendation on the nuclear dataset. This indicates a unique pattern in the context of a specific topic showing that different elements of framing theory have varying levels of impact depending on the topic. This also underlines the importance of topic-sensitive approaches in misinformation detection.

\subsection{Experiment 4: Case study}

In this experiment, we conduct a case study to analyze the similarities between two articles written about the same topic where each article has a different frame and frame elements. The articles focus on the government's proposed water reforms.

The first article has a political frame: 

\textit{"There's a lot of change being proposed by the government… Fundamentally, they're considering shifting responsibility for our three waters: water supply, wastewater, and stormwater, from local government into four large entities... The government now believes that costs of between \$120 billion and \$185b will be required: between \$4 and \$6b per year on average… The proposed three waters reform program harks back to the Havelock North water contamination event in 2016… It's on this basis that the government has concluded that four entities, aggregating all the water services across the country, offer the best and quickest opportunity to achieve the desired improvements to the three-waters networks... It's too early to ask the community..."}

While the second article has a semantic frame to show their satire:

\textit{"Oh, boy! The government is proposing some exciting changes, folks. Brace yourselves because they're considering taking control of our beloved three waters. You know, the precious water supply, wastewater, and stormwater that our local government has been responsible for?... The government estimates that we'll need a mind-boggling \$120 billion to \$185 billion over the next 30 years… Well, now they want to hand it over to these big entities called Water Supply Entities. What a brilliant idea, right?... And get this – the government thinks it would be cheaper if larger entities took over the water services. Apparently, they can borrow more, with the government's backing, of course. I mean, who needs small, local councils when you can have these big entities making all the decisions for you?..."}

The similarities calculated in Table~\ref{tab_single_similarity} of 0.86 indicate that the articles are highly similar as they both share details about the reform. However, once the frame elements are considered, the article similarity decreases to 0.61.

The political frame is informative and objective, presenting a detailed overview using formal language, and statistics to support its claims, while the semantic frame is emotional and opinionated using colloquial language and employing vivid imagery to engage readers emotionally which may risk oversimplification and bias.

To determine the classification without frame elements, our model only encodes the news article. In comparison, for classifying with the frame elements, the elements and news articles are encoded independently with their embeddings concatenated into one vector prior to classification. The inclusion of these extra features enhances the model’s performance.

For example, given the problem definition for the political frame:

\textit{``The proposed shift of responsibility for three waters from local government to four large entities known as water supply entities.”}

As well as the problem definition for the semantic frame:

\textit{``The proposed government takeover of three waters”}

The problem definition of each article highlights their differences in framing, the politically framed article’s problem definition is detailed with a neutral tone, while the semantic frame’s problem definition is short with a negative perspective. 

Both articles are classified as information without frame elements, however, once the frame elements are considered, the semantic frame is correctly classified as misinformation. This suggests that including the frame elements in our model contributes to the successful classification of misinformation by reinforcing the differences between the two semantically similar articles.

\subsection{Discussions}
We proposed a Frame Element-based Model (FEM) to distinguish misinformation from information. Several experiments are conducted providing crucial insights into the importance of elements of framing theory in misinformation detection. Results are evaluated by comparing them with baseline models. Furthermore, we analyzed the contribution of each element demonstrating the different roles of the elements. By comparing the article similarities and element similarities, we obtained insights into how the elements improve the performance of detecting misinformation. Based on the experimental results, we have the following insights observed:
\begin{itemize}
\item The results of Experiment 1 on all datasets consistently proved that incorporating the elements of framing theory while detecting misinformation stemming from portrayed facts under different narratives can help improve the performance. 

\item The parameter analysis experiment revealed that the absence of certain framing elements, particularly Problem Definition and Moral Evaluation, leads to a significant decrease in the model’s accuracy and precision. This underscores the critical role these elements play in the accurate detection of misinformation.

\item Experiment 2 also demonstrates a finding that the specific element plays a different role on different topics highlighting the potential impact of elements and underscoring the necessity for topic-sensitive approaches in misinformation detection, as different elements of framing theory have varying levels of impact depending on the topic.

\item The similarity comparison experiment further illustrated how misinformation closely mirrors authentic information in terms of framing. The results indicated an inverse relationship between similarity and model performance: lower similarity between the information and misinformation articles led to higher performance in detecting misinformation. 

\item The case study shows that not applying the framing theory to the articles can lead to the result of the article with a semantic frame incorrectly classified as information increasing the potential for misleading interpretations. It also shows that while articles may be semantically similar, the choice of framing can greatly impact the narrative of content being misleading or misinterpreted.

\end{itemize} 

\section{Conclusion and Future Work}\label{sec:conclusion}

In this work, we introduce the Framed Element-based Model (FEM) to identify misinformation in the context of news articles incorporating the elements of framing theory, i.e., Problem Definition, Causal Interpretation, Moral Evaluation, and Treatment Recommendation. This model leverages ChatGPT and deep neural networks to detect misinformation originating from accurately portrayed facts under different frames. The efficacy of FEM is demonstrated through comprehensive performance comparisons with other methods, highlighting the effectiveness of Framed Element-based approach against traditional misinformation detection models. The contribution of each element is also evaluated and analyzed along with the similarities under different conditions, indicating the importance of the specific element and showcasing how the narrative of an article is framed.

This work has laid a foundational understanding of how elements of framing theory influence the perception and interpretation of information. Building upon the insights obtained, there are several future directions. Future studies can delve into the impact of specific elements across various topics, such as the frame of treatment recommendation on the Nuclear dataset shows more impact than on the other three datasets which raises questions that can be explored in the future. Besides the contribution of each element we explored, relations among these elements also can be explored as a future direction which can help understand and enhance the ability to detect misinformation under more complex scopes.

\section*{Acknowledgment}
This project has been supported by the AUT DCT Faculty Contestable Research Fund 2023. We acknowledge the support and resources provided by AUT and thank all the participants and collaborators who dedicated their time and effort to this work.

\bibliographystyle{unsrt}  
\bibliography{references}

\end{document}